\documentclass[10pt,twocolumn,letterpaper]{article}

\usepackage{cvpr}
\usepackage{times}
\usepackage{epsfig}
\usepackage{graphicx}
\usepackage{amsmath}
\usepackage{amssymb}

\usepackage[utf8]{inputenc} % allow utf-8 input
\usepackage[T1]{fontenc}    % use 8-bit T1 fonts
\usepackage{url}            % simple URL typesetting
\usepackage{booktabs}       % professional-quality tables
\usepackage{amsfonts}       % blackboard math symbols
\usepackage{nicefrac}       % compact symbols for 1/2, etc.
\usepackage{microtype}      % microtypography
\usepackage{multirow}
\usepackage{blindtext}
\usepackage{algorithm,algpseudocode}
\usepackage{subcaption}
\usepackage{wrapfig}
\usepackage{flushend}
\newcommand{\argmax}{\mathop{\mathrm{arg\ max}}}

\usepackage[pagebackref=true,breaklinks=true,letterpaper=true,colorlinks,bookmarks=false]{hyperref}

\cvprfinalcopy % *** Uncomment this line for the final submission

\ifcvprfinal\fi
\begin{document}

%%%%%%%%% TITLE
\title{Progressive Ensemble Networks for Zero-Shot Recognition}

\author{Meng Ye	\\
Computer and Information Sciences\\ 
Temple University,
Philadelphia, USA\\
{\tt\small meng.ye@temple.edu}
\and
Yuhong Guo\\
School of Computer Science\\
Carleton University, 
Ottawa, Canada\\
{\tt\small yuhong.guo@carleton.ca}
}

\maketitle

%-----------------------------------------------------------------------------------
% Abstract
%-----------------------------------------------------------------------------------
\begin{abstract}

Despite the advancement of supervised image recognition algorithms, 
their dependence 
on the availability of labeled data and the rapid expansion of image categories 
raise the significant challenge of zero-shot learning. 
Zero-shot learning (ZSL) 
aims to transfer knowledge from labeled classes 
into unlabeled classes to reduce human labeling effort. 
In this paper, we propose a novel progressive ensemble network model 
with multiple projected label embeddings 
to address zero-shot image recognition. 
The ensemble network is built by 
learning multiple image classification functions with a shared feature extraction network 
but different label embedding representations, 
which enhance the diversity of the classifiers and facilitate 
information transfer to unlabeled classes.  
A progressive training framework is then deployed to gradually
label the most confident images in each unlabeled class with predicted pseudo-labels
and update the ensemble network with the training data augmented by the pseudo-labels. 
The proposed model performs training on both labeled and unlabeled data.
It can naturally bridge the domain shift problem in visual appearances
and be extended to the generalized zero-shot learning scenario.
We conduct experiments on multiple 
	ZSL datasets and the empirical results 
demonstrate the efficacy of the proposed model.
\end{abstract}

%-----------------------------------------------------------------------------------
% Introduction
%-----------------------------------------------------------------------------------
\section{Introduction}
Despite the effectiveness of deep convolutional neural networks (CNNs) on supervised image classification problems, zero shot learning (ZSL) remains a challenging and fundamental problem due to 
the rapid expansion of image categories and the lacking in labeled training data. 
As a special unsupervised domain adaptation, 
ZSL aims to transfer information from the source domain, a set of training classes with labeled data,
to make predictions in the target domain, a set of test classes with only unlabeled data. 
Different from standard domain adaptation, 
in ZSL the labeled training classes
and unlabeled test classes have no overlaps -- they are entirely disjoint.
Based on the visibility of the instance labels, the training classes and the test classes
are usually referred to as \emph{seen} and \emph{unseen} classes respectively. 

Existing zero-shot image recognitions have centered on deploying label embeddings
in a common semantic space, e.g., in terms of high level visual attributes, 
to bridge the domain gap between \emph{seen} and \emph{unseen} classes. 
For example, animals share some common characteristics such as `black', `yellow', `spots', `stripes' and so on. 
Thus each animal class, either seen or unseen, can be represented as a binary vector 
in the semantic attribute space,
with each element denoting the appearance/absence of certain attribute.
Much ZSL effort in this direction has focused on developing effective mapping models from the input 
visual feature space to the semantic label embedding space 
~\cite{romera2015embarrassingly,frome2013devise,bucher2016improving,kodirov2017semantic}, 
or learning suitable compatibility functions 
between the two spaces~\cite{akata2015evaluation,socher2013zero,xian2016latent}, 
to facilitate prediction information transfer from the seen classes to the unseen classes.
However, these methods identify visual-semantic mappings only on 
the labeled seen class data, which 
poses a fundamental {\em domain shift} problem 
due to the appearance variations of visual attributes across 
\emph{seen} and \emph{unseen} classes,
and has negative impact on cross-class generalization (i.e., ZSL performance) 
\cite{fu2015transductive,kodirov2015unsupervised}. 

In this paper, we propose a novel ZSL framework with 
an progressive ensemble network 
to address the domain shift problem and improve the generalization ability of ZSL.
Existing ZSL works rely on a single set of label embeddings
to build inter-class label relations for knowledge transfer, 
which can hardly to be suitable for all the unseen classes. 
Instead we construct a deep ensemble network that consists of 
multiple image classification functions with a shared feature extraction convolutional neural network 
and different label embedding representations. 
Each label embedding representation facilitates 
information transfer from the seen classes to a subset of unseen classes,
while enhancing the diversity of the multiple classifiers.
By exploiting multiple classifiers in an ensemble manner, we expect
the ensemble network can overcome the prediction noise and class bias in the original label embeddings
to gain robust zero-shot predictions.
Moreover, we exploit the unlabeled data from unseen classes
in a progressive ensemble framework to overcome the domain shift problem.
In each iteration, we select the most confidently predicted unlabeled instances
from each unseen class under the current ensemble network, 
and combine these selected instances and their predicted pseudo-labels 
with the original labeled seen class data together to refine the ensemble network parameters, 
especially its feature extraction component. 
By incorporating the unseen class instances into the ensemble network training
and dynamically refine the selected instances in each iteration,
we expect the dynamic progressive training process can effectively 
avoid the issue of overfitting to the seen classes
and improve the generalization ability of the ensemble network on unseen classes. 
With the ensemble network directly handling multi-class classification over all classes,
the proposed approach can be conveniently extended to address generalized ZSL.
We conduct experiments on three standard ZSL datasets 
under both conventional ZSL and generalized ZSL settings.
The empirical results demonstrate the proposed approach outperforms 
the state-of-the-art ZSL methods.

%-----------------------------------------------------------------------------------
% Related Work
%-----------------------------------------------------------------------------------
\section{Related Work}

%-----------------------------------------------------------------------------------
\subsection{Zero-Shot Learning}
Deploying label embeddings in a common semantic space, e.g., visual attributes, 
to bridge the gap between seen and unseen classes is the key of ZSL. 
Existing ZSL methods have mostly centered on learning 
a transferable mapping function between the input visual feature space 
and the semantic label space.
ALE~\cite{akata2013label} and DeViSE~\cite{frome2013devise} both use a linear projection to map visual features into 
the semantic space. 
LatEm~\cite{xian2016latent} uses non-linear compatibility functions to match the two spaces,
while some other works learn bilinear compatibility functions 
~\cite{akata2015evaluation,romera2015embarrassingly}. 
Neural networks are used in~\cite{zhang2017learning,lei2015predicting} to embed semantic information,
while Semantic Auto-Encoders (SAE) with reconstruction loss is used in~\cite{tsai2017learning,kodirov2017semantic} to learn better projections to the semantic space. SynC~\cite{changpinyo2016synthesized} and ConSE~\cite{norouzi2013zero} embed unseen instances as a linear combination of seen class embeddings.

Despite the differences in embedding techniques, 
these methods are trained only on seen classes and have no clue about the 
visual appearance variations in unseen classes.
They suffer from the aforementioned domain shift problem. 
Some most recent advances try to solve ZSL in a generative style. 
The work in \cite{changpinyo2017predicting} uses a linear projection to map an unseen semantic attribute vector into 
a visual feature space, which can be used for generating instances of the unseen classes. 
The work of \cite{bucher2017generating} uses a generative moment matching network to generate unseen class instances, on which a classifier is directly trained for classification. 
In~\cite{zhu2017imagine} the authors used a GAN to synthesize visual features from noisy texts. 
However the generated features in these works are not guaranteed to align well with the true unseen visual features,
and can still suffer from the domain shift problem.

%-----------------------------------------------------------------------------------
\subsection{Transductive Zero-Shot Learning}
Different from the standard zero-shot learning setting where unlabeled instances from unseen classes 
are treated as inaccessible in the training phase, 
transductive ZSL refers to the setting that unseen class instances are available during training.
As none of the unseen class instances are labeled, this setting does not violate the `zero-shot' principle. 
The existing transductive ZSL works have improved 
standard ZSL by exploiting the unseen class instances 
to overcome the domain shift problem.
In~\cite{fu2015multiview} the authors adopted a two-step procedure. They first used CCA to project both visual feature and class prototypes into a multi-view embedding space, 
and then used test instances to build a hypergraph in the embedded space 
for label propagation.
The authors of~\cite{kodirov2015unsupervised} proposed to solve ZSL from the viewpoint of unsupervised domain adaptation with sparse coding. 
In~\cite{guo2016transductive} the authors proposed to learn a shared model space on seen and unseen data
to facilitate knowledge transfer between classes. 
The work in \cite{tsai2017learning} uses auto-encoders 
to learn joint embeddings of visual and semantic vectors. 
It exploits unseen class instances to minimize a prediction loss for better adaptation. 
The work in \cite{guo2017zero} proposes to assign pseudo-labels to test instances and train embedding matrix on both seen and unseen class data. It nevertheless uses a single projection matrix to project visual features into the semantic space. 
More recently, the authors of~\cite{verma2017simple} proposed to learn generative models to predict data distribution of seen and unseen classes from their attribute vectors, and used unlabeled test data to refine the distribution parameters of target classes. 
The work in~\cite{song2018transductive} trains an end-to-end network that optimizes the loss on both seen class data and unseen test data, by minimizing the Quasi-Fully Supervised Learning loss, 
which uses target class data to reduce seen/unseen bias of the model during training.

Our proposed work belongs to transductive zero-shot learning, but differs from the existing
transductive ZSL works in two major aspects: (1) 
Instead of using one set of label embeddings that are not optimized for any target unseen class, 
our ensemble network uses 
multiple sets of label embeddings, 
each of which  
is produced by enhancing
the inter-label relations between the seen classes and a subset of unseen classes.
An ensemble combination of multiple classification functions with different output representations
can facilitate robust knowledge transfer to all the unseen classes.
(2) We use a progressive ZSL framework that dynamically
incorporates a subset of unlabeled instances selected from the unseen classes 
and their predicted pseudo-labels to gradually improve the ensemble network and prevent domain shift. 
In each iteration,
with our dynamic instance selection procedure, new instances can be selected and previous ones might be dropped, 
which provides the ability to `correct' potential bad predictions in previous iterations. 
%-----------------------------------------------------------------------------------
\subsection{Progressive Training with Pseudo-Labels}
Exploiting unlabeled data by assigning them predicted pseudo-labels in a 
static progressive training procedure
has been deployed in standard classification settings in the literature. 
A notable example is the well-known 
co-training method~\cite{blum1998combining}, which
uses two different classifiers to produce pseudo-labels on unlabelled data. 
Sharing similar ideas with co-training, a recent Tri-training method~\cite{zhou2005tri} 
also exploited outputs of three different classifiers. 
In~\cite{saito2017asymmetric}, the authors applied tri-training in solving unsupervised domain adaptation problems. 
In \cite{bengio2009curriculum}, progressive curriculum learning is used to train a model on "easy-to-hard" samples with a pre-defined scheme.
The self-paced co-training work in \cite{ma2017self} uses a progressive "easy-to-hard" strategy as well as two views of the data for training. 
The authors of \cite{wu2018exploit} proposed a progressive sampling scheme for video retrieval task.
Distinct from these works above, our proposed work proposes 
a novel ensemble network that contains multiple classification functions with different label embeddings
to address a more challenging zero-shot learning problem using a progressive procedure.

%-----------------------------------------------------------------------------------
\begin{figure*}[thbp!]
\vskip 0.05in
\centering
\includegraphics[width=0.90\textwidth]{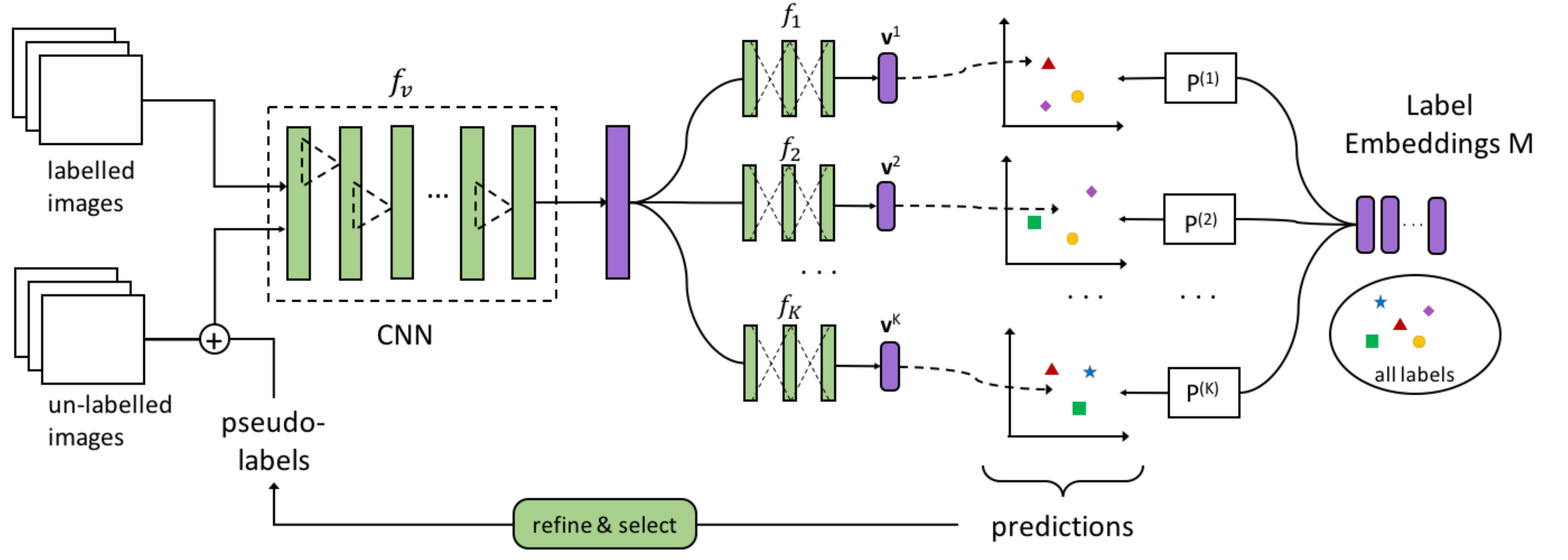}
\caption{The deep architecture of the proposed progressive ensemble network. 
The ensemble network consists of multiple ($K$) image classification functions, 
each of which is a composition of a shared image feature extraction function $f_v$ and 
an individual semantic embedding function $f_k$, i.e., $f_k\circ f_v(x)$,
with $k\in\{1,\cdots,K\}$.
We use the ResNet-34 \protect\cite{he2016deep} as the feature extraction convolutional neural network $f_v$
and use a multilayer perceptron with two hidden layers (512 units) and ReLU 
activation functions as each embedding function $f_k$.
The progressive training procedure iteratively and dynamically selects unlabeled instances
and their predicted pseudo-labels to augment the training data and refine the ensemble network.
}
\label{fig:model}
\vskip -0.1in
\end{figure*}

%-----------------------------------------------------------------------------------
% Approach
%-----------------------------------------------------------------------------------
\section{Approach}

We consider zero-shot image recognition in the following setting. 
We have a set of $N^s$ labeled images, 
$\mathcal{D}_{s}=\{(x_i, y_i)\}_{i=1}^{N^s}$, 
from $L^s$ seen classes $\mathcal{S}=\{1,\cdots,L^s\}$
such that $y_i\in\mathcal{S}$.
We also have a set of $N^u$ images,
$\mathcal{D}_{u}=\{(x_j, y_j)\}_{j=1}^{N^u}$, 
from $L^u$ unseen classes 
$\mathcal{U}=\{L^s+1,\cdots,L\}$ such that $L=L^s+L^u$,
where the labels, $\{y_j\in\mathcal{U}\}$, 
are unavailable during training. 
We aim to transfer information from the labeled data to 
predict the labels of the unlabeled instances. 
To bridge the gap between seen and unseen classes,
we also assume we have a semantic label representation matrix $M\in\mathbb{R}^{m\times L}$,
e.g., semantic attribute vectors,
for all the $L$ seen and unseen classes.  

In this section, we present a novel progressive ensemble network model for zero-shot image recognition. 
The proposed end-to-end framework is depicted in Figure~\ref{fig:model}.
It consists of multiple image classification functions with a shared feature extraction network
but different label embedding representations. 
A progressive training framework is deployed to iteratively
refine the overall ensemble network by incorporating unlabeled instances with their 
predicted pseudo-labels in a dynamic and ensemble manner. 
%
%
%-----------------------------------------------------------------------------------
\subsection{Ensemble Networks}
Following the standard ZSL scheme, we can use a convolutional neural network (CNN)
$f_v$ to extract high level visual features from an image $x$,
and then use an embedding network $f$ to embed the visual features $f_v(x)$
into the semantic space, e.g., the attribute space, of label embeddings $\mathbb{R}^m$.
Here the overall deep network $f\circ f_v(x)$ (``$\circ$" denotes a composition operation)  
forms an image classification function
for all classes, $\mathcal{S}\cup\mathcal{U}$,
which can categorize an image $x$ to the nearest class in the semantic label embedding space $\mathbb{R}^m$.
However, though a semantic label embedding matrix $M$
can enable zero-shot information transfer from the seen classes to the unseen classes,
the effectiveness of such information transfer can vary substantially 
for different unseen classes due to their various association levels with
the seen classes in the given semantic label embedding space.
It is hard to optimize the semantic associations between the seen classes 
and all unseen classes simultaneously 
with one fixed label embedding matrix $M$.
Hence we propose to project the label embeddings $M$ into $K$
different embedding spaces, 
$\{\mathcal{P}_k: \mathbb{R}^m \rightarrow \mathbb{R}^h\;| k=1,\cdots,K\}$,
to induce $K$ sets of 
different label embeddings $\{\mathcal{P}_k(M)\}$ to facilitate information transfer
to the unseen classes. 
For each label embedding matrix  $\mathcal{P}_k(M)$, we can produce
an embedding network $f_k$, e.g., a multilayer perceptron, to map $f_v(x)$ into the corresponding label
embedding space, which forms a zero-shot classification function $f_k\circ f_v(x)$.
By employing the $K$ classification functions in an ensemble manner
we expect the overall ensemble network 
can effectively reduce the impact of noise and class bias of the original 
label embeddings $M$ to
produce robust zero-shot image recognitions. 

\subsubsection{Label Embedding Projection}

We aim to use different label embeddings to capture different label associations
between seen and unseen classes. Towards this goal,
we perform the $k$-th label embedding projection $\mathcal{P}_k$ adaptively
by maximizing the weighted 
similarity score between the seen classes, $\mathcal{S}$,
and a randomly selected subset of the unseen classes, $\mathcal{Z}_k\subset\mathcal{U}$,
in the projected label embedding space.
In particular, we assume a linear projection function  $\mathcal{P}_k(M) = P^{(k)}M$,
where the projection matrix 
$P^{(k)}\in\mathbb{R}^{h\times m}$ %($h<m$) 
has orthogonal rows, i.e., $P^{(k)}P^{(k)\top}=I$.
We formulate the label embedding projection as the following maximization problem:
\begin{align}
\max_{P^{(k)}} &
\sum_{i\in \mathcal{S}, j\in \mathcal{Z}_k} 
	\mbox{tr}(P^{(k)} M_{:i}A_{ij}M_{:j}^\top P^{(k)\top})
\\
\text{subject to} &
 \quad P^{(k)}P^{(k)^\top}=I
\nonumber
\end{align}
where $M_{:i}$ denotes the $i$-th column of matrix $M$ and tr$(\cdot)$ denotes a trace function,
the association weight $A_{ij}$ is defined as the cosine similarity 
between the corresponding $i$-th and $j$-th classes in the original label representation space. 
This maximization problem has a closed-form solution:
\begin{equation}
P^{(k)} = [\mathbf{u}_1, \mathbf{u}_2, ..., \mathbf{u}_h]^\top
\label{eq:proj}
\end{equation}
where $\{\mathbf{u}_i\}_{i=1}^h$ are the top $h$ eigenvectors 
of matrix 
$\sum_{i\in \mathcal{S}, j\in \mathcal{Z}_k} \frac{1}{2}(M_{:i}A_{ij}M_{:j}^\top + M_{:j}A_{ij}M_{:i}^\top)$.

We can produce $K$ different label embedding projection matrices $\{P^{(k)}\}_{k=1}^K$
by randomly selecting $K$ different subsets of unseen classes, $\{\mathcal{Z}_k\}_{k=1}^K$. 
Each resulting label embedding matrix $P^{(k)}M$ 
encodes a different knowledge transfer structure between the seen and unseen classes. 
%%

%-----------------------------------------------------------------------------------
\subsubsection{Loss Function of the Ensemble Network }
Given labeled training instances $\mathcal{D}_{train} =\{(x_i,y_i)\}_{i=1}^N$,
the deep ensemble neural network with $K$ classification functions,
$\{f_k\circ f_v(x)\}_{k=1}^K$, can be trained by minimizing the following negative log-likelihood loss function:
\begin{align}
& \mathcal{L}(\omega_v, \omega_1, ..., \omega_K) = 
 \frac{1}{N}\sum_{i=1}^N\sum_{k=1}^K \ell_k(\mathbf{v}_i^k, y_i) 
\label{eq:objective}
\end{align}
where $(\omega_v, \omega_1, ..., \omega_K)$ denote the model parameters, 
$\mathbf{v}_i^k = f_k\circ f_v(x_i)$ denotes 
the $k$-th classifier's prediction vector of instance $x_i$ in its label embedding space, and 
$\ell_k(\cdot,\cdot)$ is a negative log-likelihood loss function computed over the 
softmax prediction scores of the $k$-th classifier:
\begin{align}
\ell_k(\mathbf{v}_i^k, c) &= -\mathrm{log}\ p_k(c|\mathbf{v}_i^k) 
\nonumber\\
&= -\mathrm{log}\frac{\text{exp}(\mathbf{v}_i^{k\top} P^{(k)}M_{:c})}
{\sum_{c'\in \mathcal{S}\cup\mathcal{U}}\text{exp}(\mathbf{v}_i^{k\top} P^{(k)}M_{:c'})}
\end{align}
%%%%%%%%
%
Note the softmax function above is defined {\em over both seen and unseen classes}. 
It is designed to include training instances from both seen and unseen classes.
Hence, although initially the labeled training data only contain 
the labeled instances from the seen classes,
such that $\mathcal{D}_{train}=\mathcal{D}_s$ and $N=N^s$, 
we will expand it to include pseudo-labeled set from unseen classes through progressive training below. 
%

%-----------------------------------------------------------------------------------
\subsubsection{Ensemble Zero-Shot Prediction}
With the multiple classification functions learned in the ensemble network, 
we can integrate the $K$ classification functions to perform zero-shot prediction
on each unlabeled instance $x_i$ from unseen classes.
We first make predictions using each of the $K$ classifiers 
based on similarity scores:
\begin{equation}
\hat{y}_i^{(k)} = \argmax_{c\in\mathcal{Z}_k}\quad \langle f_k\circ f_v(x_i),\ P^{(k)}M_{:c}\rangle
\label{eq:score}
\end{equation}
where $\langle\cdot,\cdot\rangle$ denotes the inner product of two vectors. 
As the $k$-th set of label embeddings are produced by maximizing the label associations
of seen classes and the subset of unseen classes $\mathcal{Z}_k$, 
we hence only use the $k$-th classifier for zero-shot predictions on the subset of
unseen classes $\mathcal{Z}_k$. 
Then we ensemble all the $K$ predictions to determine the predicted class
using a normalized majority voting strategy:
\begin{align}
\hat{y}_i &= \argmax_{c}\quad \phi(x_i,c) 
\label{eq:aggregate}
\\
\mbox{where}\quad &
\phi(x_i,c)  = 
\frac{\sum_{k=1}^K \mathbb{I}{[c=\hat{y}_i^{(k)}]}}{\sum_{k=1}^K \mathbb{I}{[c\in\mathcal{Z}_k]}}
\label{eq:avgscore}
\end{align}
and $\mathbb{I}[\cdot]$ denotes an indicator function that returns value 1 
when the given condition is true. 

In the case of generalized ZSL, where a test instance $x_i$ can be from either a seen 
or an unseen class,
we still compute the voting score of $x_i$ belonging to an unseen class $c$ using
the normalized voting score in Eq.(\ref{eq:avgscore}), 
but we compute the voting score of $x_i$ belonging to a seen class $c$
as its average prediction score on this class by all the $K$ classifiers, i.e., 
$\phi(x_i,c) = \frac{1}{K}\sum_{k=1}^K\langle f_k\circ f_v(x_i),\ P^{(k)}M_{:c}\rangle$.

%-----------------------------------------------------------------------------------
\subsection{Progressive Ensemble Networks}
Training with only labeled instances from the seen classes
can suffer from the aforementioned domain shift problem. 
Meanwhile our ensemble network provides a natural foundation 
for making voting-based predictions on the unseen class instances 
and incorporating pseudo-labeled instances from the unseen classes in the training process.
We hence propose to deploy a progressive training procedure
that iteratively and dynamically exploits pseudo-labeled unseen class instances 
to refine the ensemble network initially 
trained on the labeled data from seen classes, $\mathcal{D}_s$. 

The progressive training algorithm is summarized in Algorithm~\ref{algo}.
In each iteration, it uses the current ensemble network to 
predict the pseudo-label $\hat{y}_i$ with Equation (\ref{eq:aggregate}) for each unlabeled instance $x_i$ from the unseen classes.
Then for each unseen class $c\in\mathcal{U}$, it selects the top $N_{pseudo}$ instances 
with the largest prediction scores $\phi(x_i,c)$.
The instances selected from all the unseen classes together with their predicted labels 
form a pseudo-set $\mathcal{D}_{pseudo}=\{(x_i,\hat{y}_i)\}_{i=1}^{N_p}$.
The ensemble network parameters are then refined by minimizing a loss function in Equation (\ref{eq:objective})
over an augmented training set $\mathcal{D}_{train} = \mathcal{D}_{s} \cup \mathcal{D}_{pseudo}$. 
As the augmented training set contains data from both the seen classes and unseen classes,
we expect the refined ensemble network 
can overcome the domain shift problem in terms of visual appearances of semantic features
and improve zero-shot prediction performance. 
Moreover, instead of gradually increasing the pseudo-set, we dynamically update this set in each iteration
with the progressively improved ensemble network to correct potential label mistakes
in the previous pseudo-set.

\begin{algorithm}[tb]
   \caption{Progressive Training of Ensemble Networks}
   \label{algo}
\begin{algorithmic}
   \State {\bfseries Input:} labeled data from seen classes $\mathcal{D}_s$,
 unlabeled data from unseen classes $\mathcal{D}_u$, 
and label embedding matrix $M$.
   \State {\bf Initialization: }\\ 
\quad    $\mathcal{D}_{train}\gets\mathcal{D}_s$, $\mathcal{D}_{pseudo}\gets\emptyset$;
\\
\quad perform label embedding projection for $\{P^{(k)}\}_{k=1}^K$;
\\
\quad  train an end-to-end deep ensemble network on $\mathcal{D}_{train}$. 
   \Repeat
   \State predict pseudo-labels of $\mathcal{D}_u$ by Eq.(\ref{eq:aggregate}) and (\ref{eq:avgscore});
   \State generate a pseudo-labeled set $\mathcal{D}_{pseudo}$ by selecting the 
\\\qquad\quad top $N_{pseudo}$ instances from each unseen class;
   \State update the training set:  $\mathcal{D}_{train}\gets\mathcal{D}_s\cup\mathcal{D}_{pseudo}$; 
   \State refine the ensemble network parameters on $\mathcal{D}_{train}$. 
   \Until {$\mathtt{MaxIter}$}
\end{algorithmic}
\end{algorithm}

%-----------------------------------------------------------------------------------
%\input{experiments}
%-----------------------------------------------------------------------
\section{Experiments}
To investigate the empirical performance of our proposed approach, 
we conducted experiments under both conventional ZSL and generalized ZSL settings.
In this section, we present our experimental results and discussions.

\subsection{Experiment Settings}
\subsubsection{\bf Datasets} 
We used three widely used ZSL datasets with label attribute vectors to conduct experiments.
The first one is the Caltech-UCSD-Birds 200-2011 (CUB) dataset~\cite{WahCUB_200_2011}. 
It is a fine-grained dataset of bird species, containing 11,788 images of birds from 200 different species. 
Each image is also annotated with 312 attributes. 
The second one is the SUN dataset~\cite{patterson2014sun}, which contains 14,340 images from 717 different scenes. 
In this dataset each image is annotated with 102 attributes. 
The third dataset is the Animal with Attributes 2 (AWA2) dataset~\cite{xian2017zero}, 
which is an updated version of the previous AWA~\cite{lampert2009learning} dataset. 
AWA2 consists of 37,322 images from 50 animal classes. 
It also provides 85 numerical attribute values for each class. 
We used AWA2 instead of AWA as the raw image data of AWA is not publicly available any more.
Following previous ZSL works, we extracted the label embedding matrix $M$
from the attribute vectors. 

%-----------------------------------------------------------------------------------
\subsubsection{\bf Seen/Unseen Splits} 
In order to perform ZSL, a dataset needs to be split into two disjoint subsets, 
the seen classes $\mathcal{S}$ 
and the unseen classes $\mathcal{U}$. 
To perform scientific ZSL study and maintain the 
`zero-shot' principle, 
a ZSL model should never have access to the true label information of the unseen class instances
during the training phase. 
However
many ZSL approaches have used CNN models pre-trained on the ImageNet~\cite{ILSVRC} for image feature extraction.
If the pre-trained ImageNet classes have overlaps with the ZSL test classes, 
it should be considered as violating the `zero-shot' rule. 
As pointed out in the comprehensive evaluation study~\cite{xian2017zero}, 
standard splits (SS) on the ZSL datasets have 
unseen class overlaps with the 1K classes of ImageNet, 
which can lead to superior performance on these classes. 
Therefore in this study we also used the ZSL splits proposed in~\cite{xian2017zero} (PS), 
which has the same number of test classes as the SS splits but
ensures no class in ImageNet appears in the test set of ZSL. 
For the SUN dataset, except for the split with 72 test classes, 
there is another split with 10 test classes from~\cite{jayaraman2014zero}, 
which is also used in some previous works. 
We denote this split as SUN10 and the split with 72 test classes as SUN72. 
The overview of these datasets and seen/unseen class splits are summarized in Table~\ref{tbl:datasets}. 

\begin{table}[t]
\caption{Summary of three attribute datasets for ZSL.}
\label{tbl:datasets}
\begin{center}
\begin{small}
\begin{sc}
\begin{tabular}{lcccc}
\toprule
Dataset & Images & Avg. & Classes & Attr. \\
\midrule
CUB		& 11788	& $\sim$60 	& 200 (150+50)	& 312 \\
SUN		& 14340	& $\sim$20 	& 717 (645+72)	& 102 \\
AWA2		& 37322	& $\sim$750	& 50 (40+10)		& 85 \\
\bottomrule
\end{tabular}
\end{sc}
\end{small}
\end{center}
\vskip -0.1in
\end{table}

%----------------------
\begin{table*}[t!]
\caption{Conventional ZSL results. $\dagger$ denotes numbers cited from \protect\cite{xian2017zero}. 
Methods in the top part of the table reported the Top-1 accuracy results (TOP-1), while those in the bottom part reported the multi-class accuracy (MACC) results.
Numbers in bracket denote results on AWA instead of AWA2.
Best results are shown in {\bf bold} font and second best in {\it italic} font.	
	}
\setlength{\tabcolsep}{8pt}
\label{tbl:zsl}
\begin{center}
\begin{small}
\begin{sc}
\begin{tabular}{llcccccccc}
\toprule
&\multirow{2}{*}{Methods}  &\multicolumn{2}{c}{CUB} & \multicolumn{2}{c}{AWA2} & \multicolumn{2}{c}{SUN72} & SUN10\\
& & SS & PS & SS & PS & SS & PS &  \\
\midrule
\multirow{11}{*}{Top-1} &DeViSE~\cite{frome2013devise}$\dagger$			& 53.2 & 52.0 & 68.6 & 59.7 & 57.5 & 56.5 & - \\
&SynC~\cite{changpinyo2016synthesized}$\dagger$	& 54.1 & 55.6 & 71.2 & 46.6 & 59.1 & 56.3 & - \\
&ALE~\cite{akata2013label}$\dagger$				& 53.2 & 54.9 & 80.3 & 62.5 & 59.1 & 58.1 & - \\
&SJE~\cite{akata2015evaluation}$\dagger$		& 55.3 & 53.9 & 69.5 & 61.9 & 57.1 & 53.7 & - \\
&SAE~\cite{kodirov2017semantic}$\dagger$		& 33.4 & 33.3 & 80.7 & 54.1 & 42.4 & 40.3 & - \\
&ReViSE~\cite{tsai2017learning}				& 65.4 & - & (93.4) & - & - & - & - \\
	&GFZSL~\cite{verma2017simple}               & 63.8 & -    & ({\it 94.3}) &  -   &  -   &  -   & {\bf 87.0} \\
&QFSL~\cite{song2018transductive}			& {\bf 69.7} & {\bf 72.1} & 84.8 & {\bf 79.7} & {\it 61.7} & 58.3 & - \\
&Progressive Training								& 54.0 & 49.8 & 73.2 & 57.8 & 49.6 & 47.9 & 76.6 \\
&PrEN$_{w/o Proj}$					& 64.7 & 61.4 & 88.5 & 66.6 & 61.1 & {\it 60.1} & 84.4 \\
&PrEN (proposed)      					& {\it 66.9} & {\it 66.4} & {\bf 95.7} & {\it 74.1} & {\bf 63.3} & {\bf 62.9} & {\it 86.3}\\
\midrule
\multirow{5}{*}{mAcc} 
&UDA~\cite{kodirov2015unsupervised}	& 40.6 & - & (75.6) & - & - & - & - \\
&DCL~\cite{guo2017zero}				&   -  & - & (81.9) & - & - & - & 84.4 \\
&Progressive Training					    & 53.7 & 49.9 & 73.0 & 53.8 & 49.5 & 48.0 & 76.7 \\
&PrEN$_{w/o Proj}$		& {\it 64.4} & {\it 61.4} & {\it 89.4} & {\it 65.7} & {\it 61.0} & {\it 60.2} & {\it 84.5} \\
&PrEN (proposed)      			& {\bf 66.6} & {\bf 66.4} & {\bf 96.1} & {\bf 78.6} & {\bf 63.2} & {\bf 62.8} & {\bf 86.4}\\
\bottomrule
\end{tabular}
\end{sc}
\end{small}
\end{center}
\vskip -0.1in
\end{table*}

%-----------------------------------------------------------------------------------
\subsubsection{\bf Evaluation Metric} 
We adopted the popularly used Top-1 accuracy to evaluate the ZSL prediction performance. 
The Top-1 accuracy counts the proportion of correctly labeled instances in each test class and then 
takes an average over all these classes. 
To compare with some literature works, we also reported the multi-class classification accuracy results when needed.

%-----------------------------------------------------------------------------------
\subsubsection{\bf Implementation Details}
For an input image, we resized it to $224\times224$ and fed it to ResNet-34~\cite{he2016deep}. 
The 512 dimensional vector from the last average pooling layer of ResNet is 
used as visual features of the image. The ResNet is initialized by the pre-trained model on ImageNet. 
We used multilayer perceptrons (MLPs)
with two hidden layers (each with size 512) and one output layer (with size $h$) as 
the consequent embedding functions. 
ReLU activation
is applied after each layer. 
We used Adam~\cite{kingma2014adam} to train our model, with the default parameter setting $\beta_1=0.9$, $\beta_2=0.999$ and 
learning rate $\eta = 0.001$. 
We set $\mathtt{MaxIter}=20$. 
In each iteration the model is trained with 100 batches with batch size 64.
For the progressive training procedure, 
we used $N_{pseudo}=\mathtt{min}(\rho N_{avg}, N_{max})$, 
where $N_{avg}$ is the average number of images in each training class, 
$\rho$ and $N_{max}$ are set to 0.25 and 20 respectively.
We used $K=50$ different label embeddings (i.e., the number of classifiers). 
For $\{\mathcal{Z}_k\}_{k=1}^K$, which are the randomly selected subsets of unseen classes
for producing the label embedding projection matrices, we set the size of each $\mathcal{Z}_k$
as half of the unseen class number. 
We projected the original label embeddings to a lower dimension space
such $h<m$. We used h=70 in the experiments if not specifically noted.

%-----------------------------------------------------------------------------------
% Results
%-----------------------------------------------------------------------------------
%-----------------------------------------------------------------------------------
\subsection{Conventional ZSL Results}
\subsubsection{\bf Comparison Methods}
We compared the proposed Progressive Ensemble Network (PrEN) model 
with a number of state-of-the-art ZSL methods. 
These methods can be divided into two groups: 
DeViSE~\cite{frome2013devise}, SynC~\cite{changpinyo2016synthesized}, ALE~\cite{akata2013label},
SJE~\cite{akata2015evaluation}, and SAE~\cite{kodirov2017semantic}  
belong to inductive ZSL methods,
while the transductive methods include UDA~\cite{kodirov2015unsupervised}, DCL~\cite{guo2017zero}, 
ReViSE~\cite{tsai2017learning},
GFZSL~\cite{verma2017simple}, and QFSL~\cite{song2018transductive}. 
All the comparison methods used the standard fixed splits. 
We hence take the convenience to cite the results from~\cite{xian2017zero} and the literature for fair comparisons.

In order to separate the impact of the progressive training principle from our proposed ensemble framework,
we also compared with a {\em Progressive Training} baseline variant of the proposed model, 
which drops the ensemble framework to use only one classifier with the original label embeddings. 
Moreover, to investigate the effectiveness of multiple adaptive label embedding projections, 
we also tested another {\em ensemble baseline} variant, which deviates from the proposed model only 
by using the same original label embeddings for the $K$ classifiers without any projection.
We denote this baseline variant as {\em PrEN$_{w/o Proj}$}.

%-----------------------------------------------------------------------------------
\subsubsection{\bf Result Analysis}
We summarized the comparison results in Table~\ref{tbl:zsl}. 
As two different evaluation metrics, Top-1 Accuracy and multi-class accuracy,
are used in the comparison works,  
we divide the table into two parts, where the top part presents Top-1 accuracy results 
and the bottom part presents multi-class accuracy results. 
We reported the results of our proposed PrEN method in terms of both evaluation metrics. 
From the table, we notice that 
most results under `PS' are worse than their counterpart results under `SS', especially on the AWA2 datasets. 
This indicates that the overlapping of test classes with ImageNet 1K classes did bring extra benefit in performance. 
However,
the transductive ZSL methods we found are mostly evaluated under the `SS' setting, 
we hence expect their missing results under `PS' can only be worse than the reported `SS' results. 
Moreover, the transductive works, ReViSE, GFZSL, UDA and DCL, reported results on CUB, AWA or SUN10, 
but not on AWA2 and SUN72. 
Since AWA2 is nearly a drop-in-replacement of AWA~\cite{xian2017zero}, we included their results on AWA just for reference. 

From the comparison results in terms of Top-1 accuracy, 
we can see that PrEN outperforms all the five inductive methods across all datasets. 
Comparing with the transductive methods, 
PrEN produced the second best results 
on CUB, SUN10, and the `PS' split of AWA2, where QFSL performs the best.
Nevertheless, PrEN produced the best results in all the other cases, 
the `SS' split of AWA2, both `SS' and `PS' splits of SUN72. 
In particular, on the more challenging scene classification dataset SUN72 
(more unseen classes and less training data for each class), 
PrEN achieves 63.3\% and 62.9\% on `SS' and `PS' splits,
and outperforms all the other methods with notable performance gains.  
In terms of multi-class accuracy, 
our proposed PrEN largely outperforms the two transductive ZSL methods, UDA and DCL. 
For example, PrEN achieves 66.6\% on CUB and 86.4\% on SUN10, which are much better than
the 40.6\% reported by UDA on CUB and the 84.4\% reported by DCL on SUN10 respectively.
These results demonstrate the efficacy of the proposed approach for conventional
zero-shot image recognition tasks.

By comparing the proposed PrEN with the baseline variants,
we notice there are large performance gaps between
the proposed full model PrEN and the 
two variants, Progressive-Training baseline variant and PrEN$_{w/o Proj}$.
Without the ensemble architecture and the diverse label embeddings, 
the progressive training procedure alone cannot produce any effective model.
Even by just dropping the label embedding projection 
but maintaining the ensemble architecture,
PrEN$_{w/o Proj}$ still yields much inferior performance than PrEN.
These suggest 
that the proposed ensemble network architecture with the essential
label embedding projections 
forms a solid and critical foundation for incorporating pseudo-labels through 
progressive training.

%%

%-----------------------------------------------------------------------------------
\begin{table*}[t]
\caption{Generalized ZSL results in terms of average Top-1 accuracy. $\dagger$ denotes numbers cited from 
\protect\cite{xian2017zero}. `u' and `s' denotes Top-1 accuracies on unseen and seen classes, respectively. `H' denotes the harmonic mean of them.}
\vskip -.35in
\setlength{\tabcolsep}{10pt}
\begin{center}
\begin{small}
\begin{sc}
\begin{tabular}{lccc|ccc|ccc}
\toprule
\multirow{2}{*}{Methods} &\multicolumn{3}{c}{CUB} & \multicolumn{3}{c}{AWA2} & \multicolumn{3}{c}{SUN72}\\
 & u & s & H & u & s & H & u & s & H\\
\midrule
DeViSE~\cite{frome2013devise}$\dagger$			& 23.8 & 53.0 & 32.8 & 17.1 & 74.7 & { 27.8} & 16.9 & 27.4 & 20.9\\
SynC~\cite{changpinyo2016synthesized}$\dagger$	& 11.5 & 70.9 & 19.8 & 10.0 & 90.5 & 18.0 &  7.9 & 43.3 & 13.4\\
ALE~\cite{akata2013label}$\dagger$				& 23.7 & 62.8 & {34.4} & 16.8 & 76.1 & 27.5 & 21.8 & 33.1 & { 26.3}\\
SJE~\cite{akata2015evaluation}$\dagger$			& 23.5 & 59.2 & 33.6 &  8.0 & 73.9 & 14.4 & 14.7 & 30.5 & 19.8\\
	SAE~\cite{kodirov2017semantic}$\dagger$			&  7.8 & 54.0 & 13.6 &  1.1 & 82.2 &  2.2 &  8.8 & 18.0 & 11.8\\[.5ex]
%midrule
PrEN (proposed) 			            & 35.2 & 55.8 & {\bf 43.1} & 32.4 & 88.6 & {\bf 47.4} & 35.4 & 27.2 & {\bf 30.8} \\
\bottomrule
\end{tabular}
\end{sc}
\end{small}
\end{center}
\vskip -0.2in
\label{tbl:gzsl}
\end{table*}

%-----------------------------------------------------------------------------------
\paragraph{Empirical Computational Complexity.}
From above, we can see that tremendous performance gain has been achieved by 
the proposed PrEN model over its baseline Progressive Training variant. 
As PrEN involves training multiple classifiers, $K=50$ in our experiments,
while the Progressive Training variant has $K=1$, 
a natural question to ask is that how much additional computational cost is required to yield such performance gain.
Here we use the number of floating-point operations (FLOPs), 
i.e., the total number of multiplication and addition operations,
involved in passing one image from the input of the deep network architecture to the outputs
as an empirical measure of the computational complexity induced by each deep model. 
Both models share the same ResNet34 backbone structure
which involves 3.6 billion FLOPs, 
while PrEN has 49 more MLP components than the Progressive baseline,
and each component has around 0.8 million FLOPs.
Comparing to the 3.6 billion FLOPs in the backbone ResNet,
the additional 49$\times$0.8 million $\approx$ 0.04 billion FLOPs
induced by the proposed PrEN is relatively negligible,
which however contributed to the average $16.9\%$ and $17.9\%$ 
performance gain in terms of Top-1 accuracy and multi-class accuracy respectively. 
This again validates the suitability and efficacy of the proposed ensemble architecture
with localized adaptive label embedding projections.

\subsection{Parameter Sensitivity Analysis}
\begin{figure}[t!]
    \centering
        \includegraphics[width=1.6225in,height=1.50in]{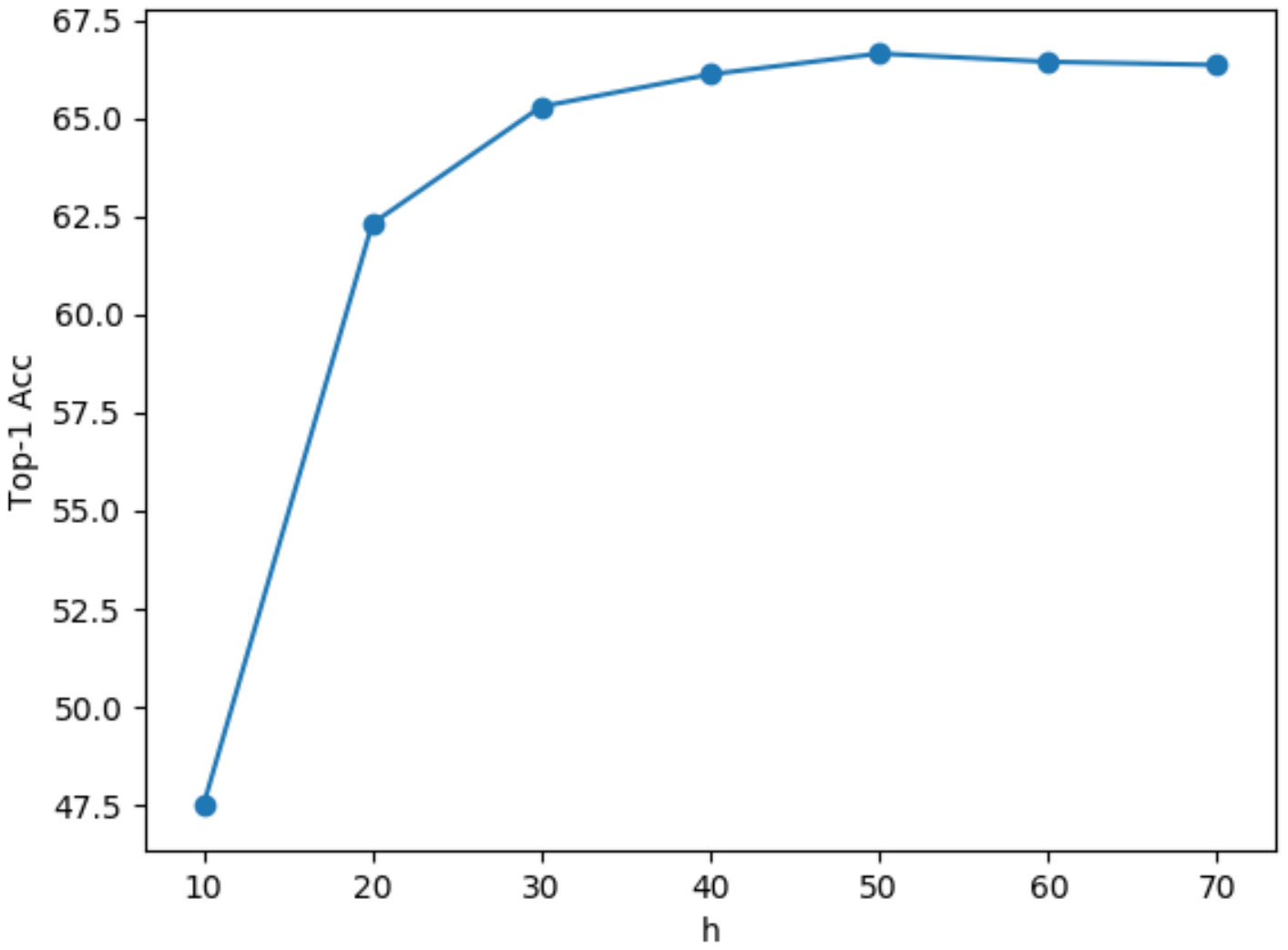}
        \includegraphics[width=1.6225in,height=1.50in]{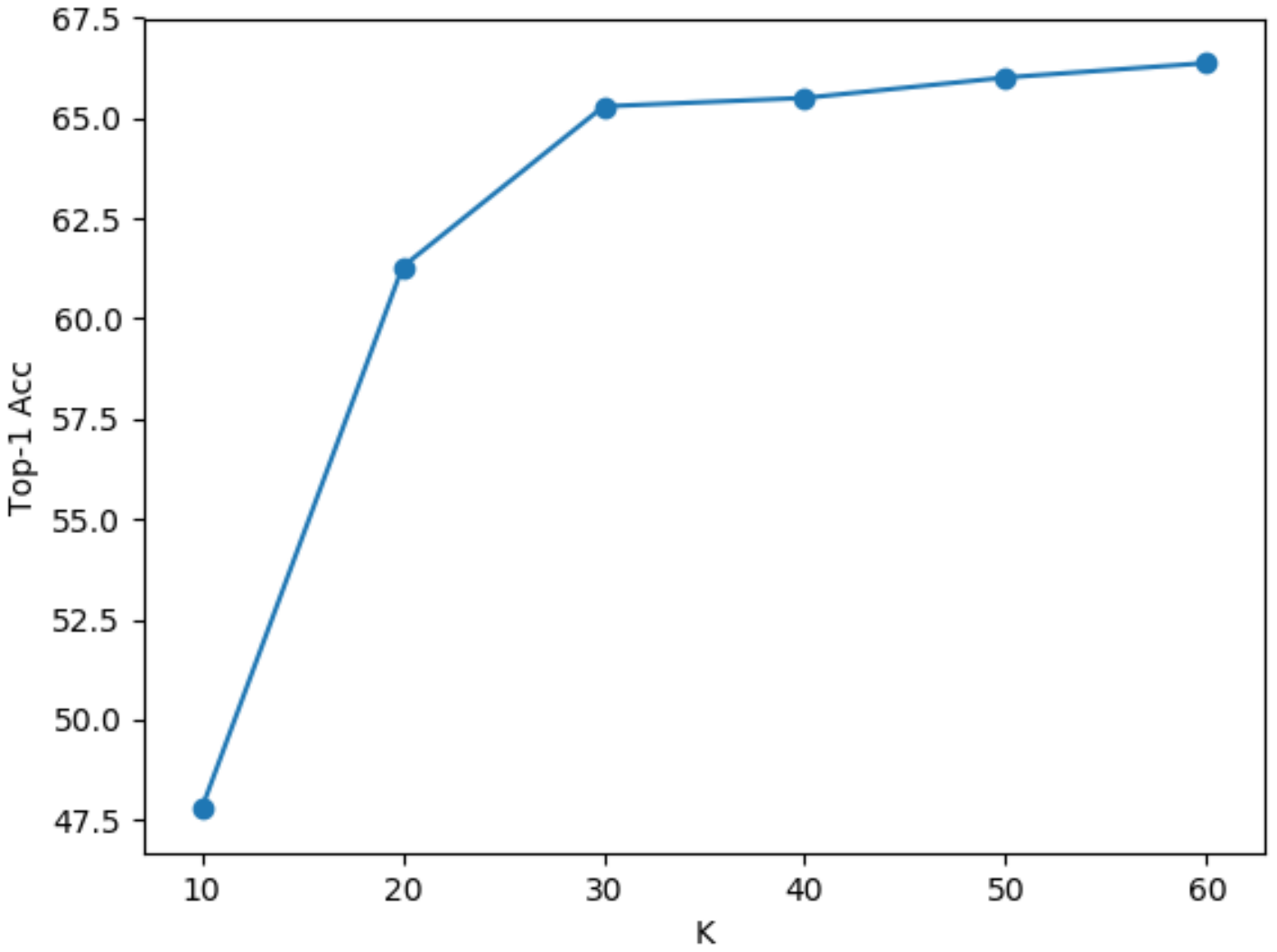}
    \caption{Parameter sensitivity analysis.}
    \label{fig:sensitivity}
\vskip -0.11in
\end{figure}

In this section, we investigate the sensitivity of the proposed model with respect to its two
hyper-parameters, $K$ and $h$.
$K$ is the number of projected label embeddings as well as the number of classifiers, 
while $h$ is the dimension of projected label embeddings.

To study how does $h$ affect the test performance, 
we performed conventional ZSL on the CUB dataset  
with the `PS' split. 
We fixed $K=50$ and 
repeated the experiment for each $h$ value
from the set $\{10, 20, 30, ..., 70\}$.
The test accuracies are reported on the left side of Figure~\ref{fig:sensitivity}.
We can see that
the test ZSL accuracies increase quickly from $h=10$ to $h=30$
and then the increase becomes very small. Nevertheless the best performance is achieved at $h=50$.
This suggests that larger dimension does help preserve useful information in the projected label embeddings. 
But even with 
a very small fraction of the original dimension, e.g., 10\%, our model
can achieve very good performance; 
on CUB a $h$ value within $(30, 70]$ would be a safe choice.

We also performed sensitivity analysis for $K$ on CUB. 
We fixed $h=70$ and 
tested different $K$ values from $\{10, 20, ..., 60\}$. 
The test accuracy results with different $K$ values
are reported
in the right subfigure of Figure~\ref{fig:sensitivity}. 
It is easy to observe that the ZSL accuracy is very poor when $K$ has a small value $10$.
Then the ZSL performance dramatically increases 
with $K$ increasing from 10 to 30, 
and the change becomes very small with $K$ further increasing to 60.
These results suggest our proposed model is not very sensitive to the hyper-parameters $K$ 
as long as it is
set to values within the reasonable range, such as $K > 30$.
%-----------------------------------------------------------------------------------
% GZSL
%-----------------------------------------------------------------------------------
\subsection{Generalized ZSL Results}
Majority of ZSL works in the literature has focused on 
the conventional ZSL setting, where the test classes are assumed to consist of only unseen classes. 
This assumption can be overly strict. 
Hence here we conducted experiments to compare the test performance of the proposed progressive ensemble network (PrEN)
with related methods under the generalized ZSL (GZSL) setting,
where the test instances can come from both seen and unseen classes.
As the classifiers within our PrEN model perform multi-class classification over all the classes, 
it can be conveniently extended to address GZSL.
For GZSL the main problem is that many unseen class instances can be wrongly classified into seen classes. 
Hence we only select pseudo instances for unseen classes 
in the first few iterations of the progressive training process,
while selecting pseudo instances for both seen and unseen classes 
in later iterations to achieve balanced performance.
To evaluate our model under GZSL, we follow the comprehensive study in~\cite{xian2017zero} to use the `PS' splits, 
and separate a random 20\% of the instances for each seen class and add these into the test set. 
We evaluated the top-1 test accuracy on unseen and seen classes separately, 
and compute their harmonic mean as the GZSL accuracy result. 
We compared to five ZSL methods that have addressed GZSL in the literature.
Although the authors of 
\cite{song2018transductive} also reported their GZSL results of 
the transductive method QFSL,  
they conducted GZSL in a non-standard and limited general setting with extra knowledge -- 
they assumed whether the unlabeled instances belong to seen vs unseen classes is known. 
For fairness,
we hence did not compare with their results. 
Our comparison results are reported in Table~\ref{tbl:gzsl}.

We can see that some comparison methods can achieve quite good performance on seen classes 
while their zero-shot accuracy on unseen classes is very low; 
for example SynC achieves 11.5\% (unseen) and 70.9\% (seen) on AWA2 , 
as well as 10.0\% (unseen) and 90.5\% (seen) on CUB. 
The overall performance of the comparison methods on all classes, under column `H', is still poor. 
We also notice there is usually a trade-off between the performance on the seen classes and that on the unseen classes,
while the harmonic mean measures the overall performance.  
The proposed PrEN though didn't yield superior performance on seen classes,
its zero-shot prediction performance on {\em unseen} classes is much better than the other comparison methods. 
Moreover, in terms of the overall GZSL performance, we can see the proposed PrEN outperforms all the comparison methods with large margins.
This validates the effectiveness of the proposed model under GZSL setting.

%-----------------------------------------------------------------------------------
\section{Conclusion}

In this paper, we proposed a novel progressive deep ensemble network 
for transductive zero-shot image recognition. 
By integrating multiple classifiers with different label embeddings, 
the ensemble network can maintain informative knowledge transfer from seen classes to unseen classes
through adaptive inter-label relations. 
By progressively refining the ensemble network parameters with pseudo-labeled test instances, 
the training procedure can alleviate the domain shift problem and avoid overfitting
to the seen classes. 
We conducted experiments on multiple standard datasets under both conventional and generalized ZSL settings. 
The proposed model has demonstrated superior performance than the state-of-the-art comparison methods.

%-----------------------------------------------------------------------------------
% References
%-----------------------------------------------------------------------------------
{\small
\bibliographystyle{ieee}
\bibliography{paperbib}
}
\end{document}